\documentclass{article}

\usepackage{arxiv}

\usepackage[utf8]{inputenc} 
\usepackage[T1]{fontenc}    
\usepackage{hyperref}       
\usepackage{url}            
\usepackage{booktabs}       
\usepackage{amsfonts}       
\usepackage{nicefrac}       
\usepackage{microtype}      
\usepackage{lipsum}		
\usepackage{graphicx}
\usepackage{natbib}
\usepackage{doi}
\usepackage{amsmath}

\renewcommand{\vec}{\mathbf}

\title{Large-scale graph representation learning\\ with very deep GNNs and self-supervision}

\date{} 					

\author{Ravichandra Addanki\thanks{All authors contributed equally.\qquad $^\mathrm{MAG}$TK worked solely on MAG240M.\qquad $^\mathrm{PCQ}$WLSL worked solely on PCQM4M.} \\
	DeepMind\\
	\texttt{ravichandraa@deepmind.com} \\
	\And
	Peter W. Battaglia$^*$ \\
	DeepMind\\
	\texttt{peterbattaglia@deepmind.com} \\
	\And
	David Budden$^*$\\
	DeepMind\\
	\texttt{budden@deepmind.com} \\
	\And
	Andreea Deac$^*$\\
	DeepMind, Mila, Universit\'{e} de Montr\'{e}al\\
	\texttt{adeac@deepmind.com} \\
	\And
	Jonathan Godwin$^*$\\
	DeepMind\\
	\texttt{jonathangodwin@deepmind.com} \\
	\And
	Thomas Keck$^{*\mathrm{MAG}}$\\
	DeepMind\\
	\texttt{thomaskeck@deepmind.com} \\
	\And
	Wai Lok Sibon Li$^{*\mathrm{PCQ}}$\\
	DeepMind\\
	\texttt{sibon@deepmind.com} \\
	\And
	Alvaro Sanchez-Gonzalez$^*$\\
	DeepMind\\
	\texttt{alvarosg@deepmind.com} \\
	\And
	Jacklynn Stott$^*$\\
	DeepMind\\
	\texttt{jacklynnstott@deepmind.com} \\
	\AND
	Shantanu Thakoor$^*$\\
	DeepMind\\
	\texttt{thakoor@deepmind.com} \\
	\And
	Petar Veli\v{c}kovi\'{c}$^*$\\
	DeepMind\\
	\texttt{petarv@deepmind.com} \\
}

\renewcommand{\shorttitle}{Large-scale graph representation learning with very deep GNNs and self-supervision}

\hypersetup{
pdftitle={Large-scale graph representation learning with very deep GNNs and self-supervision},
pdfsubject={cs.LG},
pdfauthor={Ravichandra Addanki, Peter W. Battaglia, David Budden, Andreea Deac, Jonathan Godwin, Thomas Keck, Wai Lok Sibon Li, Alvaro Sanchez-Gonzalez, Jacklynn Stott, Shantanu Thakoor, Petar Velickovic},
pdfkeywords={OGB-LSC, MPNNs, Graph Networks, BGRL, Noisy Nodes},
}

\begin{document}
\maketitle

\begin{abstract}
Effectively and efficiently deploying graph neural networks (GNNs) at scale remains one of the most challenging aspects of graph representation learning. Many powerful solutions have only ever been validated on comparatively small datasets, often with counter-intuitive outcomes---a barrier which has been broken by the Open Graph Benchmark Large-Scale Challenge (OGB-LSC). We entered the OGB-LSC with two large-scale GNNs: a deep transductive node classifier powered by bootstrapping, and a very deep (up to 50-layer) inductive graph regressor regularised by denoising objectives. Our models achieved an award-level (top-3) performance on both the MAG240M and PCQM4M benchmarks. In doing so, we demonstrate evidence of scalable self-supervised graph representation learning, and utility of very deep GNNs---both very important open issues. Our code is publicly available at: \url{https://github.com/deepmind/deepmind-research/tree/master/ogb_lsc}.
\end{abstract}


\section{Introduction}

Effective high-dimensional representation learning necessitates properly exploiting the geometry of data \citep{bronstein2021geometric}---otherwise, it is a cursed estimation problem. Indeed, early success stories of deep learning relied on imposing strong geometric assumptions, primarily that the data lives on a grid domain; either spatial or temporal. In these two respective settings, convolutional neural networks (CNNs) \citep{lecun1998gradient} and recurrent neural networks (RNNs) \citep{hochreiter1997long} have traditionally dominated.

While both CNNs and RNNs are demonstrably powerful models, with many applications of high interest, it can be recognised that most data coming from nature cannot be natively represented on a grid. Recent years are marked with a gradual shift of attention towards models that admit a more generic class of geometric structures \citep{masci2015geodesic,velivckovic2017graph,cohen2018spherical,battaglia2018relational,de2020gauge,satorras2021n}. 

In many ways, the most generic and versatile of these models are graph neural networks (GNNs). This is due to the fact that most discrete-domain inputs can be observed as instances of a graph structure. The corresponding area of graph representaton learning \citep{hamilton2020graph} has already seen immense success across industrial and scientific disciplines. GNNs have successfully been applied for drug screening \citep{stokes2020deep}, modelling the dynamics of glass \citep{bapst2020unveiling}, web-scale social network recommendations \citep{ying2018graph} and chip design \citep{mirhoseini2020chip}.

While the above results are certainly impressive, they likely only scratch the surface of what is possible with well-tuned GNN models. Many problems of real-world interest require graph representation learning \emph{at scale}: either in terms of the amount of graphs to process, or their sizes (in terms of numbers of nodes and edges). Perhaps the clearest motivation for this comes from the Transformer family of models \citep{vaswani2017attention}. Transformers operate a self-attention mechanism over a complete graph, and can hence be observed as a specific instance of GNNs \citep{joshi2020transformers}. At very large scales of natural language data, Transformers have demonstrated significant returns with the increase in capacity, as exemplified by models such as GPT-3 \citep{brown2020language}. Transformers enjoy favourable scalability properties at the expense of their functional complexity: each node's features are updated with \emph{weighted sums} of neighbouring node features. In contrast, GNNs that rely on message passing \citep{gilmer2017neural}---passing vector signals across edges that are conditioned on \emph{both} the sender and receiver nodes---are an empirically stronger class of models, especially on tasks requiring complex reasoning \citep{velivckovic2019neural} or simulations \citep{sanchez2020learning,pfaff2020learning}.

One reason why generic message-passing GNNs have not been scaled up as widely as Transformers is the lack of appropriate datasets. Only recently has the field advanced from simple transductive benchmarks of only few thousands of nodes \citep{sen2008collective,shchur2018pitfalls,morris2020tudataset} towards larger-scale real-world and synthetic benchmarks \citep{dwivedi2020benchmarking,hu2020open}, but important issues still remain. For example, on many of these tasks, randomly-initialised GNNs \citep{velivckovic2018deep}, shallow GNNs \citep{wu2019simplifying} or simple label propagation-inspired GNNs \citep{huang2020combining} can perform near the state-of-the-art level at only a fraction of the parameters. When most bleeding-edge expressive methods are unable to improve on the above, this can often lead to controversial discussion in the community. One common example is: \emph{do we even need deep, expressive GNNs?}

Breakthroughs in deep learning research have typically been spearheaded by impactful large-scale competitions. For image recognition, the most famous example is the ImageNet Large Scale Visual Recognition Challenge (ILSVRC) \citep{russakovsky2015imagenet}. In fact, the very ``deep learning revolution'' has partly been kickstarted by the success of the AlexNet CNN model of \cite{krizhevsky2012imagenet} at the ILSVRC 2012, firmly establishing deep CNNs as the workhorse of image recognition for the forthcoming decade.

Accordingly, we have entered the recently proposed Open Graph Benchmark Large-Scale Challenge (OGB-LSC) \citep{hu2021ogb}. OGB-LSC provides graph representation learning tasks at a previously unprecedented scale---millions of nodes, billions of edges, and/or millions of graphs. Further, the tasks have been designed with immediate practical relevance in mind, and it has been verified that expressive GNNs are likely to be necessary for strong performance. Here we detail our two submitted models (for the MAG240M and PCQM4M tasks, respectively), and our empirical observations while developing them. Namely, we find that the datasets' immense scale provides a great platform for demonstrating clear outperformance of very deep GNNs \citep{godwin2021very}, as well as self-supervised GNN setups such as bootstrapping \citep{thakoor2021bootstrapped}. In doing so, we have provided meaningful evidence towards a positive resolution to the above discussion: deep and expressive GNNs are, indeed, necessary at the right level of task scale and/or complexity. Our final models have achieved award-level (top-3) ranking on both MAG240M and PCQM4M.

\section{Dataset description}

The \textbf{MAG240M-LSC} dataset is a transductive node classification dataset, based on the Microsoft Academic Graph (MAG) \citep{wang2020microsoft}. It is a heterogeneous graph containing paper, author and institution nodes, with edges representing the relations between them: paper-cites-paper, author-writes-paper, author-affiliated-with-institution. All paper nodes are endowed with 768-dimensional input features, corresponding to the RoBERTa sentence embedding \citep{liu2019roberta,reimers2019sentence} of their title and abstract. MAG240M is currently the largest-scale publicly available node classification dataset by a wide margin, at $\sim$240 million nodes and $\sim$1.8 billion edges. The aim is to classify the $\sim$1.4 million arXiv papers into their corresponding topics, according to a temporal split: papers published up to 2018 are used for training, with validation and test sets including papers from 2019 and 2020, respectively. 

The \textbf{PCQM4M-LSC} dataset is an inductive graph regression dataset based on the PubChemQC project \citep{nakata2017pubchemqc}. It consists of $\sim$4 million small molecules (described by their SMILES strings). The aim is to accelerate quantum-chemical computations: especially, to predict the HOMO-LUMO gap of each molecule. The HOMO-LUMO gap is one of the most important quantum-chemical properties, since it is related to the molecules' reactivity, photoexcitation, and charge transport. The ground-truth labels for every molecule were obtained through expensive DFT (density functional theory) calculations, which may take several hours per molecule. It is believed that machine learning models, such as GNNs over the molecular graph, may obtain useful approximations to the DFT at only a fraction of the computational cost, if provided with sufficient training data \citep{gilmer2017neural}. The molecules are split with a 80:10:10 ratio into training, validation and test sets, based on their PubChem ID.

\section{GNN Architectures}

For both of the tasks above, we rely on a common \emph{encode-process-decode} blueprint \citep{hamrick2018relational}. This implies that our input features are encoded into a latent space using node-, edge- and graph-wise \emph{encoder} functions, and latent features are decoded to node-, edge- and graph- level predictions using appropriate \emph{decoder} functions. The bulk of the computational processing is powered by a \emph{processor} network, which performs multiple graph neural network layers over the encoded latents.

To formalise this, assume that our input graph, $\mathcal{G}=(\mathcal{V},\mathcal{E})$, has node features $\vec{x}_u\in\mathbb{R}^n$, edge features $\vec{x}_{uv}\in\mathbb{R}^m$ and graph-level features $\vec{x}_\mathcal{G}\in\mathbb{R}^l$, for nodes $u, v\in\mathcal{V}$ and edges $(u, v)\in\mathcal{E}$. Our encoder functions $f_n:\mathbb{R}^n\rightarrow\mathbb{R}^k$, $f_e:\mathbb{R}^m\rightarrow\mathbb{R}^k$ and $f_g:\mathbb{R}^l\rightarrow\mathbb{R}^k$ then transform these inputs into the latent space:
\begin{equation}\label{eqn:enc}
    \vec{h}_u^{(0)} = f_n(\vec{x}_u) \qquad \vec{h}_{uv}^{(0)} = f_e(\vec{x}_{uv}) \qquad
    \vec{h}_\mathcal{G}^{(0)} = f_g(\vec{x}_\mathcal{G})
\end{equation}
Our processor network then transforms these latents over several rounds of message passing:
\begin{equation}\label{eqn:proc}
    \vec{H}^{(t+1)} = P_{t+1}(\vec{H}^{(t)}) 
\end{equation}
where $\vec{H}^{(t)} = \left(\left\{\vec{h}_u^{(t)}\right\}_{u\in\mathcal{V}}, \left\{\vec{h}_{uv}^{(t)}\right\}_{(u,v)\in\mathcal{E}}, \vec{h}_\mathcal{G}^{(t)}\right)$ contains all of the latents at a particular processing step $t\geq 0$. 

The processor network $P$ is iterated for $T$ steps, recovering final latents $\vec{H}^{(T)}$. These can then be decoded into node-, edge-, and graph-level predictions (as required), using analogous decoder functions $g_n$, $g_e$ and $g_g$:
\begin{equation}
    \vec{y}_u = g_n(\vec{h}_u^{(T)}) \qquad \vec{y}_{uv} = g_e(\vec{h}_{uv}^{(T)}) \qquad
    \vec{y}_\mathcal{G} = g_g(\vec{h}_\mathcal{G}^{(T)})
\end{equation}
We will detail the specific design of $f$, $P$ and $g$ in the following sections. Generally, $f$ and $g$ are simple MLPs, whereas we use highly expressive GNNs for $P$ in order to maximise the advantage of the large-scale datasets. Specifically, we use message passing neural networks (MPNNs) \citep{gilmer2017neural} and graph networks (GNs) \citep{battaglia2018relational}. All of our models have been implemented using the jraph library \citep{jraph2020github}.

\section{MAG240M-LSC}

\paragraph{Subsampling}

Running graph neural networks over datasets that are even a fraction of MAG240M's scale is already prone to multiple scalability issues, which necessitated either aggressive subsampling \citep{hamilton2017inductive,chen2018fastgcn,zeng2019graphsaint,zou2019layer}, graph partitioning \citep{liao2018graph,chiang2019cluster} or less expressive GNN architectures \citep{rossi2020sign,bojchevski2020scaling,yu2020scalable}.

As we would like to leverage expressive GNNs, and be able to pass messages across any partitions, we opted for the subsampling approach. Accordingly, we subsample moderately-sized patches around the nodes we wish to compute latents for, execute our GNN model over them, and use the latents in the central nodes to train or evaluate the model.

We adapt the standard GraphSAGE subsampling algorithm of \citet{hamilton2017inductive}, but make several modifications to it in order to optimise it for the specific features of MAG240M. Namely: 
\begin{itemize}
    \item We perform separate subsampling procedures across edge types. For example, an author node will separately sample a pre-specified number of papers written by that author and a pre-specified number of institutions that author is affiliated with.
    \item GraphSAGE mandates sampling an exact number of neighbours for every node, and uses sampling with replacement to achieve this even when the neighbourhood size is variable. We find this to be wasteful for smaller neighbourhoods, and hence use our pre-specified neighbour counts only as an upper bound. Denoting this upper bound as $K$, and node $u$'s original neighbourhood as $\mathcal{N}_u$, we proceed\footnote{Note that, according to the previous bullet point, $K$ and $\mathcal{N}_u$ are defined on a per-edge-type basis.} as follows:
    \begin{itemize}
        \item For nodes that have fewer neighbours of a particular type than the upper bound ($|\mathcal{N}_u| \leq K$), we simply take the entire neighbourhood, without any subsampling;
        \item For nodes that have a moderate amount of neighbours ($K < |\mathcal{N}_u| \leq 5K$) we subsample $K$ neighbours \emph{without} replacement, hence we do not wastefully duplicate nodes when the memory costs are reasonable.
        \item For all other nodes ($|\mathcal{N}_u| > 5K$), we resort to the usual GraphSAGE strategy, and sample $K$ neighbours with replacement, which doesn't require an additional row-copy of the adjacency matrix.
    \end{itemize}
    \item GraphSAGE directed the edges in the patch from the subsampled neighbours to the node which sampled them, and run their GNN for the exact same number of steps as the sampling depth. We instead modify the message passing update rule to scalably make the edges \emph{bidirectional}, which naturally allows us to run deeper GNNs over the patch. The exact way in which we performed this will be detailed in the model architecture.
\end{itemize}
Taking all of the above into account, our model's subsampling strategy proceeds, starting from paper nodes as central nodes, up to a depth of two (sufficient for institution nodes to become included). We did not observe significant benefits from sampling deeper patches. Instead, we sample significantly larger patches than the original GraphSAGE paper, to exploit the wide context available for many nodes:
\begin{itemize}
    \item[\bf Depth-0] Contains the chosen central paper node.
    \item[\bf Depth-1] We sample up to $K=40$ citing papers, $K=40$ cited papers, and up to $K=20$ authors for this paper.
    \item[\bf Depth-2] We sample according to the following strategy, for all paper and author nodes sampled at depth-1:
    \begin{itemize}
        \item[\bf Papers] Identical strategy as for depth-1 papers: up to $K=40$ cited, $K=40$ citing, $K=20$ authors.
        \item[\bf Authors] We sample up to $K=40$ written papers, and up to $K=10$ affiliations for this author.
    \end{itemize}
\end{itemize}
Overall, this inflates our maximal patch size to nearly $10,000$ nodes, which makes our \emph{patches} of a comparable size to traditional \emph{full-graph} datasets \citep{sen2008collective,shchur2018pitfalls}. Coupled with the fact that MAG240M has hundreds of millions of papers to sample these patches from, our setting enables transductive node classification at previously unexplored scale. We have found that such large patches were indeed necessary for our model's performance.

One final important remark for MAG240M subsampling concerns the existence of \emph{duplicated paper nodes}---i.e. nodes with exactly the same RoBERTa embeddings. This likely corresponds to identical papers submitted to different venues (e.g. conference, journal, arXiv). For the purposes of enriching our subsampled patches, we have combined the adjacency matrix rows and columns to ``fuse'' all versions of duplicated papers together.

\paragraph{Input preprocessing} As just described, we seek to support execution of expressive GNNs on large quantities of large-scale subsampled patches. This places further stress on the model from a computational and storage perspective. Accordingly, we found it very useful to further compress the input nodes' RoBERTa features. Our qualitative analysis demostrated that their 129-dimensional PCA projections already account for 90\% of their variance. Hence we leverage these PCA vectors as the actual input paper node features.

Further, only the paper nodes are actually provided with any features. We adopt the identical strategy from the baseline LSC scripts provided by \citet{hu2021ogb} to featurise the authors and institutions. Namely, for authors, we use the average PCA features across all papers they wrote. For institutions, we use the average features across all the authors affiliated with them. We found this to be a simple and effective strategy that performed empirically better than using structural features. This is contrary to the findings of \citet{yu2020scalable}, probably because we use a more expressive GNN.

Besides the PCA-based features, our input node features $\vec{x}_u$ also contain the one-hot representation of the node's type (paper/author/institution), the node's depth in the sampled patch (0/1/2), and a bitwise representation of the papers' publication year (zeroed out for other nodes). Lastly, and according to an increasing body of research that argues for the utility of labels in transductive node classification tasks \citep{zhu2002learning,stretcu2019graph,huang2020combining}, we use the arXiv paper labels as features \citep{wang2021bag} (zeroed out for other nodes). We make sure that the validation labels are not observed at training time, and that the central node's own label is not provided. It is possible to sample the central node at depth 2, and we make sure to mask out its label if this happens.

We also endow the patches' edges with a simple edge type feature, $\vec{x}_{uv}$. It is a 7-bit binary feature, where the first three bits indicate the one-hot type of the sampling node (paper, author or institution) and the next four bits indicate the one-hot type of the sampled neighbour (cited paper, citing paper, author or institution). We found running a standard GNN over these edge-type features more performant than running a heterogeneous GNN---once again contrary to existing baseline results \citep{hu2021ogb}, and likely because of the expressivity of our processor GNN.

\paragraph{Model architecture}

For the GNN architecture we have used on MAG240M, our encoders and decoders are both two-layer MLPs, with a hidden size of 512 features. The node and edge encoders' output layers compute 256 features, and we retain this dimensionality for $\vec{h}^{(t)}_u$ and $\vec{h}^{(t)}_{uv}$ across all steps $t$.

Our processor network is a deep message-passing neural network (MPNN) \citep{gilmer2017neural}. It computes message vectors, $\vec{m}^{(t)}_{uv}$, to be sent across the edge $(u, v)$, and then aggregates them in the receiver nodes as follows:
\begin{align}\label{eqn:mpnn1}
    \vec{m}^{(t+1)}_{uv} &= \psi_{t+1}\left(\vec{h}^{(t)}_u, \vec{h}^{(t)}_v, \vec{h}^{(0)}_{uv}\right)\\\label{eqn:mpnn2}
    \vec{h}^{(t+1)}_u &= \phi_{t+1}\left(\vec{h}^{(t)}_u, \sum_{u\in\mathcal{N}_v} \vec{m}^{(t+1)}_{vu},  \sum_{v\in\mathcal{N}_u} \vec{m}^{(t+1)}_{uv}\right) 
\end{align}
Taken together, Equations \ref{eqn:mpnn1}--\ref{eqn:mpnn2} fully specify the operations of the $P_{t+1}$ network in Equation \ref{eqn:proc}. The message function $\psi_{t+1}$ and the update function $\phi_{t+1}$ are both two-layer MLPs, with identical hidden and output sizes to the encoder network. We note two specific aspects of the chosen MPNN:
\begin{itemize}
    \item We did not find it useful to use global latents or update edge latents (Equation \ref{eqn:mpnn1} uses $\vec{h}_{uv}^{(0)}$ at all times and does not include $\vec{h}_\mathcal{G}$). This is likely due to the fact that the prediction is strongly centred at the central node, and that the edge features and types do not encode additional information.
    \item Note the third input in Equation \ref{eqn:mpnn2}, which is not usually included in MPNN formulations. In addition to pooling all incoming messages, we also pool all \emph{outgoing} messages a node sends, and concatenate that onto the input to the sender node's update function. This allowed us to simulate bidirectional edges without introducing additional scalability issues, allowing us to prototype MPNNs whose depth exceeded the subsampling depth.
\end{itemize}
The process is repeated for $T=4$ message passing layers, after which $\vec{h}_u^{(4)}$ for the central node is sent to the decoder network for predictions.

\paragraph{Bootstrapping objective} 

The non-arXiv papers within MAG240M are unlabelled and hence, under a standard node classification training regime, would contribute only implicitly to the learning algorithm (as neighbours of labelled papers). Early work on self-supervised graph representation learning \citep{velivckovic2018deep} had already shown this could be a wasteful approach, even on small-scale transductive benchmarks. Appropriately using the unlabelled nodes can provide the model with a wealth of information about the feature and network structure, which cannot be easily recovered from supervision alone. On a dataset like MAG240M---which contains $\sim$120$\times$ more unlabelled papers than labelled ones---we have been able to observe significant gains from deploying such methods.

Especially, we leverage bootstrapped graph latents (BGRL) \citep{thakoor2021bootstrapped}, a recently-proposed scalable method for self-supervised learning on graphs. Rather than contrasting several node representations across multiple views, BGRL \emph{bootstraps} the GNN to make a node's embeddings be predictive of its embeddings from another view, under a target GNN. The target network's parameters are always set to an exponential moving average (EMA) of the GNN parameters. Formally, let $f^-$ and $P^-$ be the target versions of the encoder and processor networks (periodically updated to the EMA of $f$ and $P$'s parameters), and $\vec{X}$ and $\widetilde{\vec{X}}$ be two views of an input patch (in terms of features, adjacency structure or both). Then, BGRL performs the following computations:
\begin{equation}
    \vec{H}^{(T)} = P(f(\vec{X})) \qquad
    \widetilde{\vec{H}}^{(T)} = P^-(f^-(\widetilde{\vec{X}}))
\end{equation}
where $P(f({\bf X}))$ is short-hand for applying Equation \ref{eqn:enc}, followed by repeatedly applying Equations \ref{eqn:mpnn1}--\ref{eqn:mpnn2} for $T$ steps. The BGRL loss is then optimised to make the central node embedding $\vec{h}^{(T)}_u$ predictive of its counterpart, $\widetilde{\vec{h}}^{(T)}_u$. This is done by projecting $\vec{h}^{(T)}_u$ to another representation using a \emph{projector network}, $p$, as follows:
\begin{equation}
    \vec{z}_u = p(\vec{h}_u^{(T)})
\end{equation}
where $p$ is a two-layer MLP with identical hidden and output size as our encoder MLPs. We then optimise the cosine similarity between the projector output and $\widetilde{\vec{h}}^{(T)}_u$:
\begin{equation}
    \mathcal{L}_\mathrm{BGRL} = -2\frac{\vec{z}_u^\top\widetilde{\vec{h}}_u^{(T)}}{\|\vec{z}_u\|\|\widetilde{\vec{h}}_u^{(T)}\|}
\end{equation}
using stochastic gradient ascent. Once training is completed, the projector network $p$ is discarded.

This approach, inspired by BYOL \citep{grill2020bootstrap}, eliminates the need for crafting negative samples, reduces the storage requirements of the model, and its pointwise loss aligns nicely with our patch-wise learning setting, as we can focus on performing the bootstrapping objective on each central node separately. All of this made BGRL a natural choice in our setting, and we have found that we can easily apply it at scale. 

Previously, BGRL has been applied on moderately-sized graphs with less expressive GNNs, showing modest returns. Conversely, we find the benefits of BGRL were truly demonstrated with stronger GNNs on the large-scale setting of MAG240M. Not only does BGRL monotonically improve when increasing proportions of unlabelled-to-labelled nodes during training, it consistently outperformed relevant self-supervised GNNs such as GRACE \citep{zhu2020deep}.

Ultimately, our submitted model is trained with an auxiliary BGRL objective, with each batch containing a $10:1$ ratio of unlabelled to labelled node patches. Just as in the BGRL paper, we obtain the two input patch views by applying dropout \citep{srivastava2014dropout} on the input features (with $p=0.4$) and DropEdge \citep{rong2019dropedge} (with $p=0.2$), independently on each view. The target network ($f^-$, $P^-$) parameters are updated with EMA decay rate $\epsilon=0.999$.

\paragraph{Training and regularisation} 

We train our GNN to minimise the cross-entropy for predicting the correct topic over the labelled central nodes in the training patches, added together with the BGRL objective for the unlabelled central nodes. We use the AdamW SGD optimiser \citep{loshchilov2017decoupled} with hyperparameters $\beta_1=0.9$, $\beta_2=0.999$ and weight decay rate of $\lambda=10^{-5}$. We use a cosine learning rate schedule with base learning rate $\eta=0.01$ and $50,000$ warm-up steps, decayed over $500,000$ training iterations. Optimisation is performed over dynamically-batched data: we fill up each training minibatch with sampled patches until any of the following limits are exceeded: $\sim84,000$ nodes, $\sim185,000$ edges, or $256$ patches.

To regularise our model, we perform early stopping on the accuracy over the validation dataset, and apply feature dropout (with $p=0.3$) and DropEdge \citep{rong2019dropedge} (with $p=0.25$) at every message passing layer of the GNN. We further apply layer normalisation \citep{ba2016layer} to intermediate outputs of all of our MLP modules.

\paragraph{Evaluation} At evaluation time, we make advantage of the transductive and subsampled learning setup to enhance our predictions even further: first, we make sure that the model has access to \textbf{all} validation labels as inputs at test time, as this knowledge may be highly indicative. Further, we make sure that any ``fused'' copies of duplicated nodes also provide that same label as input. As our predictions are potentially conditioned on the specific topology of the subsampled patch, for each test node we average our predictions over 50 subsampled patches---an ensembling trick which consistently improved our validation performance. Lastly, given that we already use EMA as part of BGRL's target network, for our evaluation predictions we use the EMA parameters, as they are typically slightly more stable.

\section{PCQM4M-LSC}

\paragraph{Input preprocessing}

For featurising our molecules within PCQM4M, we initially follow the baseline scripts provided by \citet{hu2021ogb} to convert SMILES strings into molecular graphs. Therein, every node is represented by a 9-dimensional feature vector, $\vec{x}_u$, including properties such as atomic number and chirality. Further, every edge is endowed with 3-dimensional features, $\vec{x}_{uv}$, including bond types and stereochemistry. Mirroring prior work with GNNs for quantum-chemical computations \citep{gilmer2017neural}, we found it beneficial to maintain graph-level features (in the form of a ``master node''), which we initialise at $\vec{x}_\mathcal{G}=\vec{0}$.

As will soon become apparent, our experiments on the PCQM4M benchmark leveraged GNNs that are substantially deeper than most previously studied GNNs for quantum-chemical tasks, or otherwise. While there is implicit expectation to compute useful ``cheap'' chemical features from the SMILES string, such as molecular fingerprints, partial charges, etc., our experiments clearly demonstrated that most of them do not meaningfully impact performance of our GNNs. This indicates that very deep GNNs are likely implicitly able to compute such features without additional guidance.

The exception to this have been \emph{conformer features}, corresponding to approximated three-dimensional coordinates of every atom. These are very expensive to obtain accurately. However, using RDKit \citep{landrum2013rdkit}, we have been able to obtain conformer estimates that allowed us to attain slightly improved performance with a (slightly) shallower GNN. Specifically, we use the experimental torsion knowledge distance geometry (ETKDGv3) algorithm \citep{wang2020improving} to recover conformers that satisfy essential geometric constraints, without violating our time limits.

Once conformers are obtained, we do not use their raw coordinates as features---these have many equivalent formulations that depend on the algorithm's initialisation. Instead, we encode their displacements (a 3-dimensional vector recording distances along each axis) and their distances (scalar norm of the displacement) as additional edge features concatenated with $\vec{x}_{uv}$. Note that RDKit's algorithm is not powerful enough to extract conformers for every molecule within PCQM4M; for about $0.1\%$ of the dataset, the returned conformers will be NaN.

Lastly, we also attempted to use more computationally intensive forms of conformer generation---including energy optimisation using the universal force field (UFF) \citep{rappe1992uff} and the Merck molecular force field (MMFF) \citep{halgren1996merck}. In both cases, we did not observe significant returns compared to using rudimentary conformers.

\paragraph{Model architecture}

For the GNN architecture we have used on PCQM4M, our encoders and decoders are both three-layer MLPs, computing 512 features in every hidden layer. The node, edge and graph-level encoders' output layers compute 512 features, and we retain this dimensionality for $\vec{h}^{(t)}_u$, $\vec{h}^{(t)}_{uv}$ and $\vec{h}^{(t)}_\mathcal{G}$ across all steps $t$.

For our processor network, we use a very deep Graph Network (GN) \citep{battaglia2018relational}. Each GN block computes updated node, edge and graph latents, performing aggregations across them whenever appropriate. Fully expanded out, the computations of one GN block can be represented as follows:
\begin{align}\label{eqn:gn1}
    \vec{h}^{(t+1)}_{uv} &= \psi_{t+1}\left(\vec{h}^{(t)}_u, \vec{h}^{(t)}_v, \vec{h}^{(t)}_{uv}, \vec{h}^{(t)}_\mathcal{G}\right)\\\label{eqn:gn2}
    \vec{h}^{(t+1)}_u &= \phi_{t+1}\left(\vec{h}^{(t)}_u, \sum_{u\in\mathcal{N}_v} \vec{h}^{(t+1)}_{vu}, \vec{h}_\mathcal{G}^{(t)}\right)\\\label{eqn:gn3}
    \vec{h}^{(t+1)}_\mathcal{G} &= \rho_{t+1}\left(\sum_{u\in\mathcal{V}}\vec{h}_u^{(t+1)}, \sum_{(u, v)\in\mathcal{E}}\vec{h}_{uv}^{(t+1)}, \vec{h}_\mathcal{G}^{(t)}\right)
\end{align}
Taken together, Equations \ref{eqn:gn1}--\ref{eqn:gn3} fully specify the operations of the $P_{t+1}$ network in Equation \ref{eqn:proc}. The edge update function $\psi_{t+1}$, node update function $\phi_{t+1}$ and graph update function $\rho_{t+1}$ are all three-layer MLPs, with identical hidden and output sizes to the encoder network.

The process is repeated for $T=32$ message passing layers, after which the computed latents $\vec{h}_u^{(32)}$, $\vec{h}_{uv}^{(32)}$ and $\vec{h}_\mathcal{G}^{(32)}$ are sent to the decoder network for relevant predictions. Specifically, the global latent vector $\vec{h}_\mathcal{G}^{(32)}$ is used to predict the molecule's HOMO-LUMO gap. Our work thus constitutes a successful application of very deep GNNs, providing evidence towards ascertaining positive utility of such models. We note that, while most prior works on GNN modelling seldom use more than eight steps of message passing \citep{brockschmidt2020gnn}, we observe monotonic improvements of deeper GNNs on this task, all the way to 32 layers when the validation performance plateaus.

\paragraph{Non-conformer model}

Recalling our prior discussion about conformer features occasionally not being trivially computable, we also trained a GN which does not exploit conformer-based features. While we observe largely the same trends, we find that they tend to allow for even deeper and wider GNNs before plateauing. Namely, our optimised non-conformer GNN computes 1,024-dimensional hidden features in every MLP, and iterates Equations \ref{eqn:gn1}--\ref{eqn:gn3} for $T=50$ message passing steps. Such a model performed marginally worse than the conformer GNN overall, while significantly improving the mean absolute error (MAE) on the $0.1\%$ of validation molecules without conformers.

\paragraph{Denoising objective}

Our very deep GNNs have, in the first instance, been enabled by careful regularisation. By far, the most impactful method for our GNN regressor on PCQM4M has been Noisy Nodes \citep{godwin2021very}, and our results largely echo the findings therein.

The main observation of Noisy Nodes is that very deep GNNs can be strongly regularised by appropriate denoising objectives. Noisy Nodes perturbs the input node or edge features in a pre-specified way, then requires the decoder to reconstruct the un-perturbed information from the GNN's latent representations.

In the case of the flat input features, we have deployed a Noisy Nodes objective on both atom types and bond types: randomly replacing each atom and each bond type with a uniformly sampled one, with probability $p=0.05$. The model then performs node/edge classification based on the final latents (e.g., $\vec{h}_u^{(32)}$, $\vec{h}_{uv}^{(32)}$ for the conformer GNN), to reconstruct the initial types. Requiring the model to correctly infer and rectify such noise is implicitly imbuing it with knowledge of chemical constraints, such as valence, and is a strong empirical regulariser. Note that, in this discrete-feature setting, Noisy Nodes can be seen as a more general case of the BERT-like objectives from \citet{hu2019strategies}. The main difference is that Noisy Nodes takes a more active role in requiring denoising---as opposed to unmasking, where it is known upfront which nodes have been noised, and the effects of noising are always predictable.

When conformers or displacements are available, a richer class of denoising objectives may be imposed on the GNN. Namely, it is possible to perturb the individual nodes' coordinates slightly, and then require the network to reconstruct the original displacement and/or distances---this time using edge regression on the output latents of the processor GNN. The Noisy Nodes manuscript had shown that, under such perturbations, it is possible to achieve state-of-the-art results on quantum chemical calculations \emph{without} requiring an explicitly equivariant architecture---only a very deep traditional GNN. Our preliminary results indicate a similar trend on the PCQM4M dataset.

\paragraph{Training and regularisation}

We train our GNN to minimise the mean absolute error (MAE) for predicting the DFT-simulated HOMO-LUMO gap based on the decoded global latent vectors. This objective is combined with any auxiliary tasks imposed by noisy nodes (e.g. cross-entropy on reconstructing atom and bond types, MAE on regressing denoised displacements). We use the Adam SGD optimiser \citep{kingma2014adam} with hyperparameters $\beta_1=0.9$, $\beta_2=0.95$. We use a cosine learning rate schedule with initial learning rate $\eta=10^{-5}$ and $50,000$ warm-up steps, peaking at $\eta=10^{-4}$, and decaying over $500,000$ training iterations. We optimise over dynamically-batched data: we fill each training minibatch until exceeding any of the following limits: $1,024$ atoms, $2,560$ bonds, or $64$ molecules.

To regularise our model, we perform early stopping on the validation MAE, and apply feature dropout \citep{srivastava2014dropout} (with $p=0.1$) and DropEdge \citep{rong2019dropedge} (with $p=0.1$) at every message passing layer.

\paragraph{Evaluation}

At evaluation time, we exploit several known facts about the HOMO-LUMO gap, and our conformer generation procedure, to achieve ``free'' reductions in MAE. 

Firstly, it is known that the HOMO-LUMO gap cannot be negative, and that it is possible for our model to make (very rare) vastly inflated predictions on validation data if it encounters an out-of-distribution molecule. We ameliorate both of these issues by clipping the network's predictions in the $[0, 20]\ \mathrm{eV}$ range.

Secondly, as discussed, for a very small fraction ($0.1\%$ of molecules), RDKit was unable to compute conformers. We found that it was useful to fall back to the 50-layer non-conformer GNN in these cases, rather than assuming a default value. The observed reductions in MAE were significant across those specific validation molecules only. 

Finally, we consistently track the exponential moving average (EMA) of our model's parameters (with decay rate $\epsilon=0.9999$), and use it for evaluation. EMA parameters are generally known to be more stable than their online counterparts, an observation that held in our case as well.

\section{Ensembling and training on validation}

Once we established the top single-model architectures for both our MAG240M and PCQM4M entries, we found it very important to perform two post-processing steps: (a) re-train on the validation set, (b) ensemble various models together.

Re-training on validation data offers a great additional wealth of learning signal, even just by the sheer volume of data available in the OGB-LSC. But aside from this, the way in which the data was split offers even further motivation. On MAG240M, for example, the temporal split implies that validation papers (from 2019) are most relevant to classifying test papers (from 2020)---simply put, because they both correspond to the latest trends in scholarship.

However, training on the full validation set comes with a potentially harmful drawback: no held-out dataset would remain to early-stop on. In a setting where overfitting can easily occur, we found the risk to vastly outweigh the rewards. Instead, we decided to randomly partition the validation data into $k=10$ equally-sized folds, and perform a cross-validation-style setup: we train $k$ different models, each one observing the training set \emph{and} $k-1$ validation folds as its training data, validating and early stopping on the held-out fold. Each model holds out a different fold, allowing us to get an overall validation estimate over the entire dataset by combining their respective predictions.

While this approach may not correspond to the intended dataset splits, we have verified that the scores on individual held-out folds match the patterns observed on models that did not observe any validation data. This gave us further reassurance that no unintended strong overfitting had happened as a result.

Another useful outcome of our $k$-fold approach is that it allowed us a very natural way to perform ensembling as well: simply aggregating all of the $k$ models' predictions would give us a certain mixture of experts, as each of the $k$ models had been trained on a slightly modified training set. Our final ensembled models employ exactly this strategy, with the inclusion of two seeds per fold. This brings our overall number of ensembled models to 20, and these ensembles correspond to our final submissions on both MAG240M and PCQM4M.

\section{Experimental evaluation}

In this section we provide experimental evidence to substantiate the various claims we have made about the key modifications in our model, hoping to advise future research on large scale graph representation learning. To eliminate any possible confounding effects of ensembling, all results reported in this section will be on a single model, evaluated on the provided validation data. We report average performance and standard deviation over three seeds.

\paragraph{MAG240M-LSC}

\begin{figure}
    \centering
    \includegraphics[width=\linewidth]{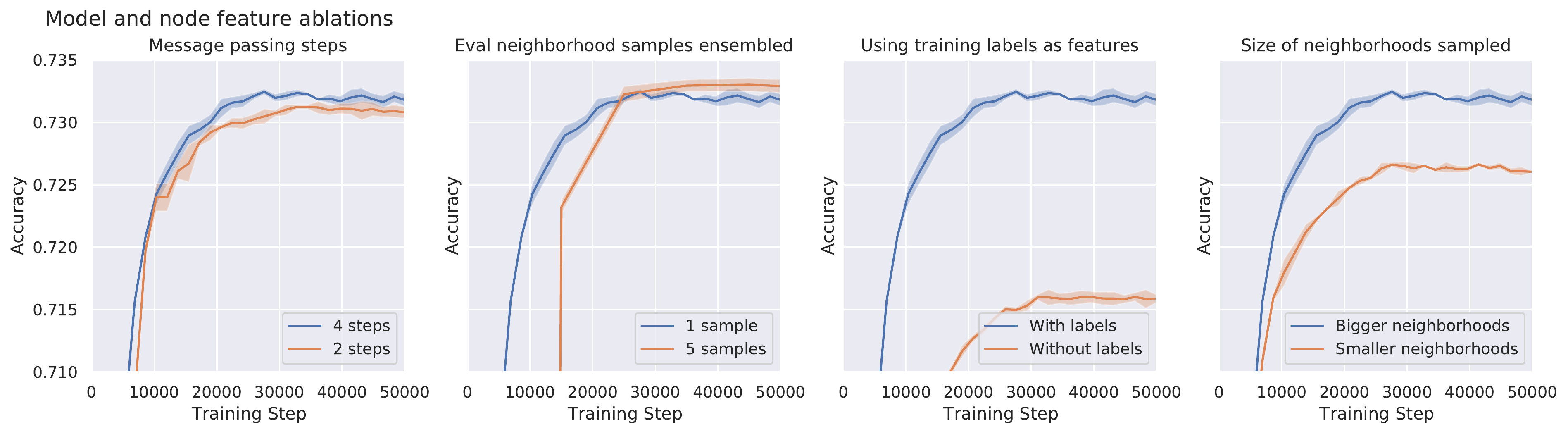}
    \caption{Ablation studies on our final MAG240M-LSC model, covering aspects of the model depth (\emph{left}), validation-time sample ensembling (\emph{middle-left}), using labels as features (\emph{middle-right}), and subsampling strategy (\emph{right}).}
    \label{fig:mag_abl_1}
\end{figure}

\begin{figure}
    \centering
    \includegraphics[width=\linewidth]{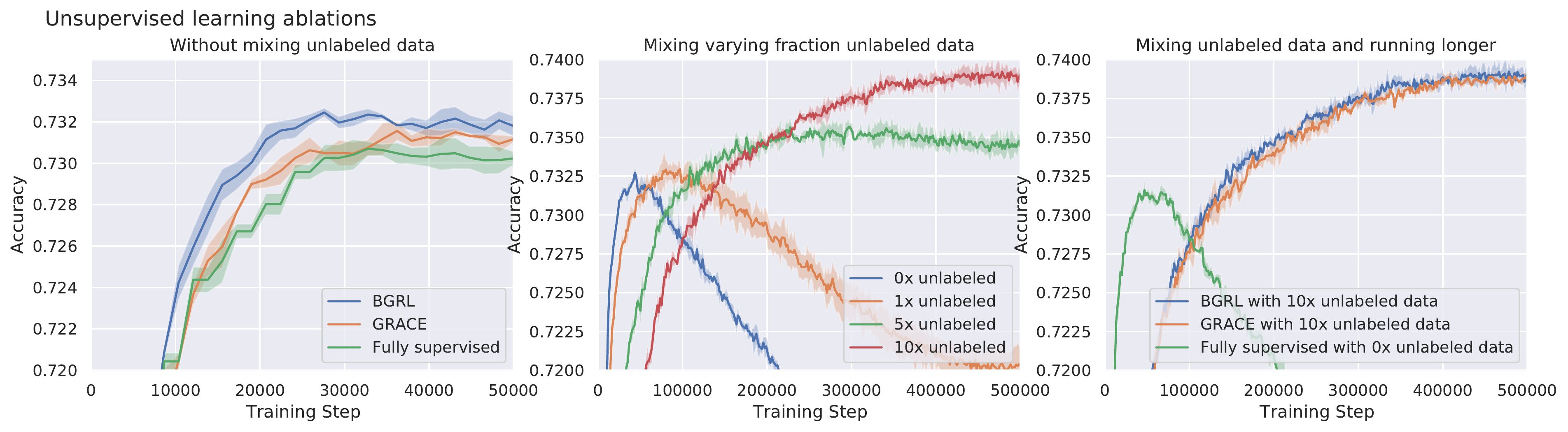}
    \caption{Ablation studies on self-supervised learning within our MAG240M-LSC entry, showing the influence of various self-supervised objectives (\emph{left}), using unlabelled nodes as targets (\emph{middle}) and running for longer (\emph{right}). Please note, the left-most plot covers training over 50,000 steps, while the other two cover training over 500,000 steps.}
    \label{fig:mag_abl_2}
\end{figure}

We will follow the plots in Figures \ref{fig:mag_abl_1}--\ref{fig:mag_abl_2}, which seek to uncover various contributing factors to our model's ultimate performance. We proceed one claim at a time.

\emph{Making networks deeper than the patch diameter can help.} We find that making the edges in every subsampled patch bidirectional allowed for doubling the message passing steps (to four) with a significant validation accuracy improvement, in spite of the fact that the MPNN was now deeper than the patch diameter. See Figure \ref{fig:mag_abl_1} (left).

\emph{Ensembling over multiple subsamples helps.} We find that averaging our network's prediction over several randomly subsampled patches at evaluation time consistently improved performance. See Figure \ref{fig:mag_abl_1} (middle-left).

\emph{Using training labels as features helps.} On transductive tasks, we confirm that using the training node label as an additional feature provides a substantial boost to validation performance, if done carefully. See Figure \ref{fig:mag_abl_1} (middle-right).

\emph{Larger patches help.} Providing the model with a larger context (by subsampling more neighbours) proved significantly helpful to our downstream performance. See Figure \ref{fig:mag_abl_1} (right).

\emph{Self-supervised objectives help---especially BGRL.} We first validate that combining a traditional cross-entropy loss with a self-supervised loss is beneficial to final performance observed. Further, we show that BGRL \citep{thakoor2021bootstrapped} can significantly outperform GRACE \citep{zhu2020deep} in the large-scale regime. See Figure \ref{fig:mag_abl_2} (left).

\emph{Self-supervised learning on unlabelled nodes helps.} One of the major promises of self-supervised learning is allowing access to a vast quantity of unlabelled nodes, which now can be used as targets. We recover consistent, monotonic gains from incorporating increasing amounts of unlabelled nodes within our training routine. See Figure \ref{fig:mag_abl_2} (middle).

\emph{Self-supervised learning allows for more robust models.} Finally, the regularising effect of self-supervised learning means that we can train our models for $10\times$ longer without suffering any overfitting effects. See Figure \ref{fig:mag_abl_2} (right).

\begin{figure}
    \centering
    \includegraphics[width=\linewidth]{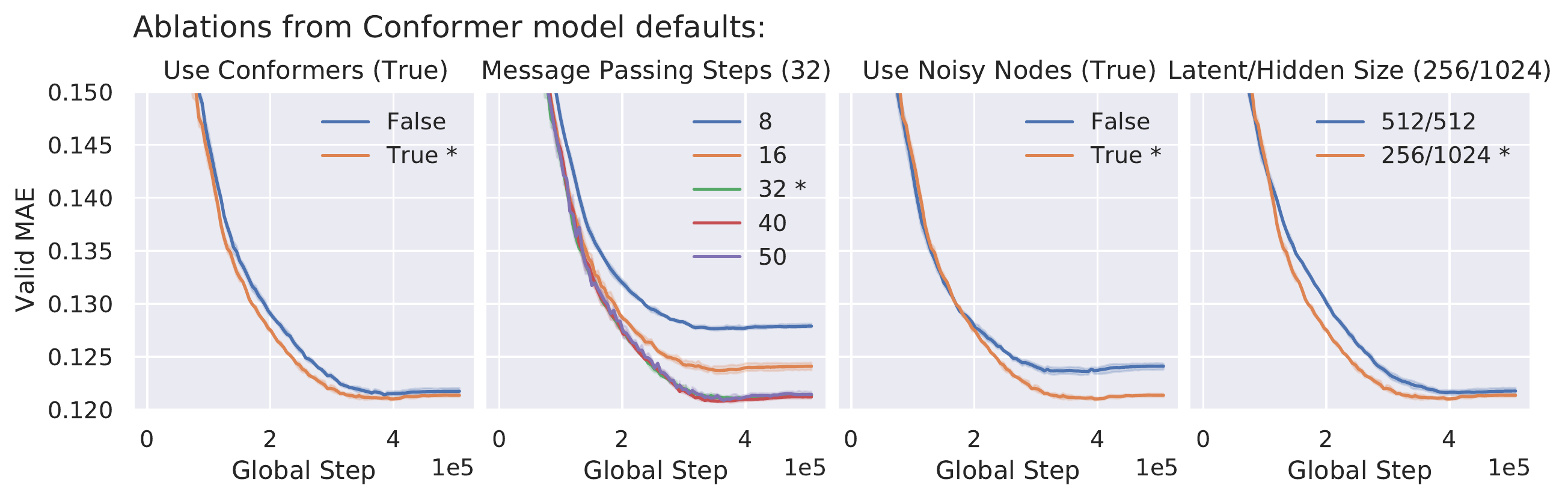}
    \includegraphics[width=\linewidth]{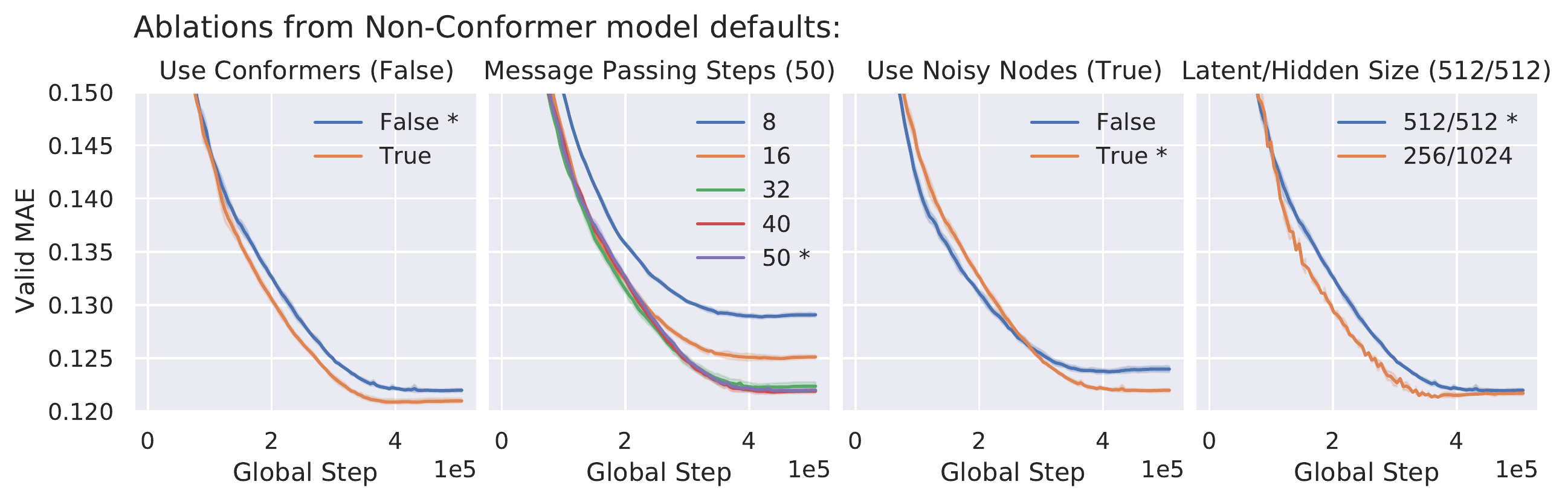}
    \caption{Ablation studies on our final PCQM4M-LSC models, covering aspects of conformer usage (\emph{left}), message passing steps (\emph{middle-left}), Noisy Nodes regularisation (\emph{middle-right}) and latent/hidden dimensions (\emph{right}). Results shown both for our conformer (\emph{top}) and non-conformer model (\emph{bottom}). * indicates our final model's chosen hyperparameter.}
    \label{fig:pcq_abl_1}
\end{figure}

\paragraph{PCQM4M-LSC} We follow Figure \ref{fig:pcq_abl_1}, which investigates key design aspects in our PCQM4M-LSC models.

\emph{Using conformer-based features helps.} Utilising features based on RDKit conformers, in the manner described before, proved beneficial to final performance. Note that the gains over our 50-layer non-conformer model are irrelevant, given that the non-conformer model is only applied over molecules where conformers cannot be computed. See Figure \ref{fig:pcq_abl_1} (top-left and bottom-left).

\emph{Deeper models help.} We demonstrate consistent, monotonic gains for larger numbers of message passing steps, at least up to 32 layers---and in the case of the non-conformer model, up to 50 layers. See Figure \ref{fig:pcq_abl_1} (top-middle-left and bottom-middle-left).

\emph{Noisy Nodes help.} Lastly, we show that the regulariser proposed in Noisy Nodes \citep{godwin2021very} proved very effective for this quantum-chemical task as well. It was the key behind the monotonic improvements of our models with depth. Note, for example, that removing Noisy Nodes from our best performing model makes its performance comparable with models that are at least twice as shallow. See Figure \ref{fig:pcq_abl_1} (top-middle-right and bottom-middle-right).

\emph{Wider message functions help.} Towards the end of the contest, we noted that performance gains are possible when favouring wider message functions (in terms of hidden size of their MLP layers) opposed to the latent size of the GNN. We subsequently noticed that such a regime (256 latent dimensions, 1,024-dimensional hidden layers) consistently improved our non-conformer model as well. See Figure \ref{fig:pcq_abl_1} (top-right and bottom-right).

\section{Results and Discussion}

Our final ensembled models achieved a \emph{validation accuracy of 77.10\%} on MAG240M, and \emph{validation MAE of 0.110} on PCQM4M. Translated on the LSC test sets, we recover \textbf{75.19\% test accuracy} on MAG240M and \textbf{0.1205 test MAE} on PCQM4M. We incur a minimal amount of distribution shift, which is a testament to our principled ensembling and post-processing strategies, in spite of using labels as inputs for MAG240M or training on validation for both tasks.

Our entries have been designated as \emph{awardees} (ranked \textbf{in the top-3}) on both MAG240M and PCQM4M, solidifying the impact that very deep expressive graph neural networks can have on large scale datasets of industrial and scientific relevance. Further, we demonstrate how several recently proposed auxiliary objectives for GNN training, such as BGRL \citep{thakoor2021bootstrapped} and Noisy Nodes \citep{godwin2021very} can both be highly impactful at the right dataset scales. We hope that our work serves towards resolving several open disputes in the community, such as the utility of very deep GNNs, and the influence of self-supervision in this setting.

In many ways, the OGB has been to graph representation learning what ImageNet has been to computer vision. We hope that OGB-LSC is only the first in a series of events designed to drive research on GNN architectures forward, and sincerely thank the OGB team for all their hard work and effort in making a contest of this scale possible and accessible.

\bibliographystyle{plainnat}
\bibliography{template}

\begin{thebibliography}{65}
\providecommand{\natexlab}[1]{#1}
\providecommand{\url}[1]{\texttt{#1}}
\expandafter\ifx\csname urlstyle\endcsname\relax
  \providecommand{\doi}[1]{doi: #1}\else
  \providecommand{\doi}{doi: \begingroup \urlstyle{rm}\Url}\fi

\bibitem[Ba et~al.(2016)Ba, Kiros, and Hinton]{ba2016layer}
Jimmy~Lei Ba, Jamie~Ryan Kiros, and Geoffrey~E Hinton.
\newblock Layer normalization.
\newblock \emph{arXiv preprint arXiv:1607.06450}, 2016.

\bibitem[Bapst et~al.(2020)Bapst, Keck, Grabska-Barwi{\'n}ska, Donner, Cubuk,
  Schoenholz, Obika, Nelson, Back, Hassabis, et~al.]{bapst2020unveiling}
Victor Bapst, Thomas Keck, A~Grabska-Barwi{\'n}ska, Craig Donner, Ekin~Dogus
  Cubuk, Samuel~S Schoenholz, Annette Obika, Alexander~WR Nelson, Trevor Back,
  Demis Hassabis, et~al.
\newblock Unveiling the predictive power of static structure in glassy systems.
\newblock \emph{Nature Physics}, 16\penalty0 (4):\penalty0 448--454, 2020.

\bibitem[Battaglia et~al.(2018)Battaglia, Hamrick, Bapst, Sanchez-Gonzalez,
  Zambaldi, Malinowski, Tacchetti, Raposo, Santoro, Faulkner,
  et~al.]{battaglia2018relational}
Peter~W Battaglia, Jessica~B Hamrick, Victor Bapst, Alvaro Sanchez-Gonzalez,
  Vinicius Zambaldi, Mateusz Malinowski, Andrea Tacchetti, David Raposo, Adam
  Santoro, Ryan Faulkner, et~al.
\newblock Relational inductive biases, deep learning, and graph networks.
\newblock \emph{arXiv preprint arXiv:1806.01261}, 2018.

\bibitem[Bojchevski et~al.(2020)Bojchevski, Klicpera, Perozzi, Kapoor, Blais,
  R{\'o}zemberczki, Lukasik, and G{\"u}nnemann]{bojchevski2020scaling}
Aleksandar Bojchevski, Johannes Klicpera, Bryan Perozzi, Amol Kapoor, Martin
  Blais, Benedek R{\'o}zemberczki, Michal Lukasik, and Stephan G{\"u}nnemann.
\newblock Scaling graph neural networks with approximate pagerank.
\newblock In \emph{Proceedings of the 26th ACM SIGKDD International Conference
  on Knowledge Discovery \& Data Mining}, pages 2464--2473, 2020.

\bibitem[Brockschmidt(2020)]{brockschmidt2020gnn}
Marc Brockschmidt.
\newblock Gnn-film: Graph neural networks with feature-wise linear modulation.
\newblock In \emph{International Conference on Machine Learning}, pages
  1144--1152. PMLR, 2020.

\bibitem[Bronstein et~al.(2021)Bronstein, Bruna, Cohen, and
  Veli{\v{c}}kovi{\'c}]{bronstein2021geometric}
Michael~M Bronstein, Joan Bruna, Taco Cohen, and Petar Veli{\v{c}}kovi{\'c}.
\newblock Geometric deep learning: Grids, groups, graphs, geodesics, and
  gauges.
\newblock \emph{arXiv preprint arXiv:2104.13478}, 2021.

\bibitem[Brown et~al.(2020)Brown, Mann, Ryder, Subbiah, Kaplan, Dhariwal,
  Neelakantan, Shyam, Sastry, Askell, et~al.]{brown2020language}
Tom~B Brown, Benjamin Mann, Nick Ryder, Melanie Subbiah, Jared Kaplan, Prafulla
  Dhariwal, Arvind Neelakantan, Pranav Shyam, Girish Sastry, Amanda Askell,
  et~al.
\newblock Language models are few-shot learners.
\newblock \emph{arXiv preprint arXiv:2005.14165}, 2020.

\bibitem[Chen et~al.(2018)Chen, Ma, and Xiao]{chen2018fastgcn}
Jie Chen, Tengfei Ma, and Cao Xiao.
\newblock Fastgcn: fast learning with graph convolutional networks via
  importance sampling.
\newblock \emph{arXiv preprint arXiv:1801.10247}, 2018.

\bibitem[Chiang et~al.(2019)Chiang, Liu, Si, Li, Bengio, and
  Hsieh]{chiang2019cluster}
Wei-Lin Chiang, Xuanqing Liu, Si~Si, Yang Li, Samy Bengio, and Cho-Jui Hsieh.
\newblock Cluster-gcn: An efficient algorithm for training deep and large graph
  convolutional networks.
\newblock In \emph{Proceedings of the 25th ACM SIGKDD International Conference
  on Knowledge Discovery \& Data Mining}, pages 257--266, 2019.

\bibitem[Cohen et~al.(2018)Cohen, Geiger, K{\"o}hler, and
  Welling]{cohen2018spherical}
Taco~S Cohen, Mario Geiger, Jonas K{\"o}hler, and Max Welling.
\newblock Spherical cnns.
\newblock \emph{arXiv preprint arXiv:1801.10130}, 2018.

\bibitem[de~Haan et~al.(2020)de~Haan, Weiler, Cohen, and Welling]{de2020gauge}
Pim de~Haan, Maurice Weiler, Taco Cohen, and Max Welling.
\newblock Gauge equivariant mesh cnns: Anisotropic convolutions on geometric
  graphs.
\newblock \emph{arXiv preprint arXiv:2003.05425}, 2020.

\bibitem[Dwivedi et~al.(2020)Dwivedi, Joshi, Laurent, Bengio, and
  Bresson]{dwivedi2020benchmarking}
Vijay~Prakash Dwivedi, Chaitanya~K Joshi, Thomas Laurent, Yoshua Bengio, and
  Xavier Bresson.
\newblock Benchmarking graph neural networks.
\newblock \emph{arXiv preprint arXiv:2003.00982}, 2020.

\bibitem[Gilmer et~al.(2017)Gilmer, Schoenholz, Riley, Vinyals, and
  Dahl]{gilmer2017neural}
Justin Gilmer, Samuel~S Schoenholz, Patrick~F Riley, Oriol Vinyals, and
  George~E Dahl.
\newblock Neural message passing for quantum chemistry.
\newblock In \emph{International Conference on Machine Learning}, pages
  1263--1272. PMLR, 2017.

\bibitem[Godwin et~al.(2020)Godwin, Keck, Battaglia, Bapst, Kipf, Li,
  Stachenfeld, Veli\v{c}kovi\'{c}, and Sanchez-Gonzalez]{jraph2020github}
Jonathan Godwin, Thomas Keck, Peter Battaglia, Victor Bapst, Thomas Kipf, Yujia
  Li, Kimberly Stachenfeld, Petar Veli\v{c}kovi\'{c}, and Alvaro
  Sanchez-Gonzalez.
\newblock {J}raph: {A} library for graph neural networks in jax., 2020.
\newblock URL \url{http://github.com/deepmind/jraph}.

\bibitem[Godwin et~al.(2021)Godwin, Schaarschmidt, Gaunt, Sanchez-Gonzalez,
  Rubanova, Veli{\v{c}}kovi{\'c}, Kirkpatrick, and Battaglia]{godwin2021very}
Jonathan Godwin, Michael Schaarschmidt, Alexander Gaunt, Alvaro
  Sanchez-Gonzalez, Yulia Rubanova, Petar Veli{\v{c}}kovi{\'c}, James
  Kirkpatrick, and Peter Battaglia.
\newblock Very deep graph neural networks via noise regularisation.
\newblock \emph{arXiv preprint arXiv:2106.07971}, 2021.

\bibitem[Grill et~al.(2020)Grill, Strub, Altch{\'e}, Tallec, Richemond,
  Buchatskaya, Doersch, Pires, Guo, Azar, et~al.]{grill2020bootstrap}
Jean-Bastien Grill, Florian Strub, Florent Altch{\'e}, Corentin Tallec,
  Pierre~H Richemond, Elena Buchatskaya, Carl Doersch, Bernardo~Avila Pires,
  Zhaohan~Daniel Guo, Mohammad~Gheshlaghi Azar, et~al.
\newblock Bootstrap your own latent: A new approach to self-supervised
  learning.
\newblock \emph{arXiv preprint arXiv:2006.07733}, 2020.

\bibitem[Halgren(1996)]{halgren1996merck}
Thomas~A Halgren.
\newblock Merck molecular force field. i. basis, form, scope, parameterization,
  and performance of mmff94.
\newblock \emph{Journal of computational chemistry}, 17\penalty0
  (5-6):\penalty0 490--519, 1996.

\bibitem[Hamilton(2020)]{hamilton2020graph}
William~L Hamilton.
\newblock Graph representation learning.
\newblock \emph{Synthesis Lectures on Artifical Intelligence and Machine
  Learning}, 14\penalty0 (3):\penalty0 1--159, 2020.

\bibitem[Hamilton et~al.(2017)Hamilton, Ying, and
  Leskovec]{hamilton2017inductive}
William~L Hamilton, Rex Ying, and Jure Leskovec.
\newblock Inductive representation learning on large graphs.
\newblock \emph{arXiv preprint arXiv:1706.02216}, 2017.

\bibitem[Hamrick et~al.(2018)Hamrick, Allen, Bapst, Zhu, McKee, Tenenbaum, and
  Battaglia]{hamrick2018relational}
Jessica~B Hamrick, Kelsey~R Allen, Victor Bapst, Tina Zhu, Kevin~R McKee,
  Joshua~B Tenenbaum, and Peter~W Battaglia.
\newblock Relational inductive bias for physical construction in humans and
  machines.
\newblock \emph{arXiv preprint arXiv:1806.01203}, 2018.

\bibitem[Hochreiter and Schmidhuber(1997)]{hochreiter1997long}
Sepp Hochreiter and J{\"u}rgen Schmidhuber.
\newblock Long short-term memory.
\newblock \emph{Neural computation}, 9\penalty0 (8):\penalty0 1735--1780, 1997.

\bibitem[Hu et~al.(2019)Hu, Liu, Gomes, Zitnik, Liang, Pande, and
  Leskovec]{hu2019strategies}
Weihua Hu, Bowen Liu, Joseph Gomes, Marinka Zitnik, Percy Liang, Vijay Pande,
  and Jure Leskovec.
\newblock Strategies for pre-training graph neural networks.
\newblock \emph{arXiv preprint arXiv:1905.12265}, 2019.

\bibitem[Hu et~al.(2020)Hu, Fey, Zitnik, Dong, Ren, Liu, Catasta, and
  Leskovec]{hu2020open}
Weihua Hu, Matthias Fey, Marinka Zitnik, Yuxiao Dong, Hongyu Ren, Bowen Liu,
  Michele Catasta, and Jure Leskovec.
\newblock Open graph benchmark: Datasets for machine learning on graphs.
\newblock \emph{arXiv preprint arXiv:2005.00687}, 2020.

\bibitem[Hu et~al.(2021)Hu, Fey, Ren, Nakata, Dong, and Leskovec]{hu2021ogb}
Weihua Hu, Matthias Fey, Hongyu Ren, Maho Nakata, Yuxiao Dong, and Jure
  Leskovec.
\newblock Ogb-lsc: A large-scale challenge for machine learning on graphs.
\newblock \emph{arXiv preprint arXiv:2103.09430}, 2021.

\bibitem[Huang et~al.(2020)Huang, He, Singh, Lim, and
  Benson]{huang2020combining}
Qian Huang, Horace He, Abhay Singh, Ser-Nam Lim, and Austin~R Benson.
\newblock Combining label propagation and simple models out-performs graph
  neural networks.
\newblock \emph{arXiv preprint arXiv:2010.13993}, 2020.

\bibitem[Joshi(2020)]{joshi2020transformers}
Chaitanya Joshi.
\newblock Transformers are graph neural networks.
\newblock \emph{The Gradient}, 2020.

\bibitem[Kingma and Ba(2014)]{kingma2014adam}
Diederik~P Kingma and Jimmy Ba.
\newblock Adam: A method for stochastic optimization.
\newblock \emph{arXiv preprint arXiv:1412.6980}, 2014.

\bibitem[Krizhevsky et~al.(2012)Krizhevsky, Sutskever, and
  Hinton]{krizhevsky2012imagenet}
Alex Krizhevsky, Ilya Sutskever, and Geoffrey~E Hinton.
\newblock Imagenet classification with deep convolutional neural networks.
\newblock \emph{Advances in neural information processing systems},
  25:\penalty0 1097--1105, 2012.

\bibitem[Landrum(2013)]{landrum2013rdkit}
Greg Landrum.
\newblock Rdkit: A software suite for cheminformatics, computational chemistry,
  and predictive modeling, 2013.

\bibitem[LeCun et~al.(1998)LeCun, Bottou, Bengio, and
  Haffner]{lecun1998gradient}
Yann LeCun, L{\'e}on Bottou, Yoshua Bengio, and Patrick Haffner.
\newblock Gradient-based learning applied to document recognition.
\newblock \emph{Proceedings of the IEEE}, 86\penalty0 (11):\penalty0
  2278--2324, 1998.

\bibitem[Liao et~al.(2018)Liao, Brockschmidt, Tarlow, Gaunt, Urtasun, and
  Zemel]{liao2018graph}
Renjie Liao, Marc Brockschmidt, Daniel Tarlow, Alexander~L Gaunt, Raquel
  Urtasun, and Richard Zemel.
\newblock Graph partition neural networks for semi-supervised classification.
\newblock \emph{arXiv preprint arXiv:1803.06272}, 2018.

\bibitem[Liu et~al.(2019)Liu, Ott, Goyal, Du, Joshi, Chen, Levy, Lewis,
  Zettlemoyer, and Stoyanov]{liu2019roberta}
Yinhan Liu, Myle Ott, Naman Goyal, Jingfei Du, Mandar Joshi, Danqi Chen, Omer
  Levy, Mike Lewis, Luke Zettlemoyer, and Veselin Stoyanov.
\newblock Roberta: A robustly optimized bert pretraining approach.
\newblock \emph{arXiv preprint arXiv:1907.11692}, 2019.

\bibitem[Loshchilov and Hutter(2017)]{loshchilov2017decoupled}
Ilya Loshchilov and Frank Hutter.
\newblock Decoupled weight decay regularization.
\newblock \emph{arXiv preprint arXiv:1711.05101}, 2017.

\bibitem[Masci et~al.(2015)Masci, Boscaini, Bronstein, and
  Vandergheynst]{masci2015geodesic}
Jonathan Masci, Davide Boscaini, Michael Bronstein, and Pierre Vandergheynst.
\newblock Geodesic convolutional neural networks on riemannian manifolds.
\newblock In \emph{Proceedings of the IEEE international conference on computer
  vision workshops}, pages 37--45, 2015.

\bibitem[Mirhoseini et~al.(2020)Mirhoseini, Goldie, Yazgan, Jiang, Songhori,
  Wang, Lee, Johnson, Pathak, Bae, et~al.]{mirhoseini2020chip}
Azalia Mirhoseini, Anna Goldie, Mustafa Yazgan, Joe Jiang, Ebrahim Songhori,
  Shen Wang, Young-Joon Lee, Eric Johnson, Omkar Pathak, Sungmin Bae, et~al.
\newblock Chip placement with deep reinforcement learning.
\newblock \emph{arXiv preprint arXiv:2004.10746}, 2020.

\bibitem[Morris et~al.(2020)Morris, Kriege, Bause, Kersting, Mutzel, and
  Neumann]{morris2020tudataset}
Christopher Morris, Nils~M Kriege, Franka Bause, Kristian Kersting, Petra
  Mutzel, and Marion Neumann.
\newblock Tudataset: A collection of benchmark datasets for learning with
  graphs.
\newblock \emph{arXiv preprint arXiv:2007.08663}, 2020.

\bibitem[Nakata and Shimazaki(2017)]{nakata2017pubchemqc}
Maho Nakata and Tomomi Shimazaki.
\newblock Pubchemqc project: a large-scale first-principles electronic
  structure database for data-driven chemistry.
\newblock \emph{Journal of chemical information and modeling}, 57\penalty0
  (6):\penalty0 1300--1308, 2017.

\bibitem[Pfaff et~al.(2020)Pfaff, Fortunato, Sanchez-Gonzalez, and
  Battaglia]{pfaff2020learning}
Tobias Pfaff, Meire Fortunato, Alvaro Sanchez-Gonzalez, and Peter~W Battaglia.
\newblock Learning mesh-based simulation with graph networks.
\newblock \emph{arXiv preprint arXiv:2010.03409}, 2020.

\bibitem[Rapp{\'e} et~al.(1992)Rapp{\'e}, Casewit, Colwell, Goddard~III, and
  Skiff]{rappe1992uff}
Anthony~K Rapp{\'e}, Carla~J Casewit, KS~Colwell, William~A Goddard~III, and
  W~Mason Skiff.
\newblock Uff, a full periodic table force field for molecular mechanics and
  molecular dynamics simulations.
\newblock \emph{Journal of the American chemical society}, 114\penalty0
  (25):\penalty0 10024--10035, 1992.

\bibitem[Reimers and Gurevych(2019)]{reimers2019sentence}
Nils Reimers and Iryna Gurevych.
\newblock Sentence-bert: Sentence embeddings using siamese bert-networks.
\newblock \emph{arXiv preprint arXiv:1908.10084}, 2019.

\bibitem[Rong et~al.(2019)Rong, Huang, Xu, and Huang]{rong2019dropedge}
Yu~Rong, Wenbing Huang, Tingyang Xu, and Junzhou Huang.
\newblock Dropedge: Towards deep graph convolutional networks on node
  classification.
\newblock \emph{arXiv preprint arXiv:1907.10903}, 2019.

\bibitem[Rossi et~al.(2020)Rossi, Frasca, Chamberlain, Eynard, Bronstein, and
  Monti]{rossi2020sign}
Emanuele Rossi, Fabrizio Frasca, Ben Chamberlain, Davide Eynard, Michael
  Bronstein, and Federico Monti.
\newblock Sign: Scalable inception graph neural networks.
\newblock \emph{arXiv preprint arXiv:2004.11198}, 2020.

\bibitem[Russakovsky et~al.(2015)Russakovsky, Deng, Su, Krause, Satheesh, Ma,
  Huang, Karpathy, Khosla, Bernstein, et~al.]{russakovsky2015imagenet}
Olga Russakovsky, Jia Deng, Hao Su, Jonathan Krause, Sanjeev Satheesh, Sean Ma,
  Zhiheng Huang, Andrej Karpathy, Aditya Khosla, Michael Bernstein, et~al.
\newblock Imagenet large scale visual recognition challenge.
\newblock \emph{International journal of computer vision}, 115\penalty0
  (3):\penalty0 211--252, 2015.

\bibitem[Sanchez-Gonzalez et~al.(2020)Sanchez-Gonzalez, Godwin, Pfaff, Ying,
  Leskovec, and Battaglia]{sanchez2020learning}
Alvaro Sanchez-Gonzalez, Jonathan Godwin, Tobias Pfaff, Rex Ying, Jure
  Leskovec, and Peter Battaglia.
\newblock Learning to simulate complex physics with graph networks.
\newblock In \emph{International Conference on Machine Learning}, pages
  8459--8468. PMLR, 2020.

\bibitem[Satorras et~al.(2021)Satorras, Hoogeboom, and Welling]{satorras2021n}
Victor~Garcia Satorras, Emiel Hoogeboom, and Max Welling.
\newblock E (n) equivariant graph neural networks.
\newblock \emph{arXiv preprint arXiv:2102.09844}, 2021.

\bibitem[Sen et~al.(2008)Sen, Namata, Bilgic, Getoor, Galligher, and
  Eliassi-Rad]{sen2008collective}
Prithviraj Sen, Galileo Namata, Mustafa Bilgic, Lise Getoor, Brian Galligher,
  and Tina Eliassi-Rad.
\newblock Collective classification in network data.
\newblock \emph{AI magazine}, 29\penalty0 (3):\penalty0 93--93, 2008.

\bibitem[Shchur et~al.(2018)Shchur, Mumme, Bojchevski, and
  G{\"u}nnemann]{shchur2018pitfalls}
Oleksandr Shchur, Maximilian Mumme, Aleksandar Bojchevski, and Stephan
  G{\"u}nnemann.
\newblock Pitfalls of graph neural network evaluation.
\newblock \emph{arXiv preprint arXiv:1811.05868}, 2018.

\bibitem[Srivastava et~al.(2014)Srivastava, Hinton, Krizhevsky, Sutskever, and
  Salakhutdinov]{srivastava2014dropout}
Nitish Srivastava, Geoffrey Hinton, Alex Krizhevsky, Ilya Sutskever, and Ruslan
  Salakhutdinov.
\newblock Dropout: a simple way to prevent neural networks from overfitting.
\newblock \emph{The journal of machine learning research}, 15\penalty0
  (1):\penalty0 1929--1958, 2014.

\bibitem[Stokes et~al.(2020)Stokes, Yang, Swanson, Jin, Cubillos-Ruiz, Donghia,
  MacNair, French, Carfrae, Bloom-Ackermann, et~al.]{stokes2020deep}
Jonathan~M Stokes, Kevin Yang, Kyle Swanson, Wengong Jin, Andres Cubillos-Ruiz,
  Nina~M Donghia, Craig~R MacNair, Shawn French, Lindsey~A Carfrae, Zohar
  Bloom-Ackermann, et~al.
\newblock A deep learning approach to antibiotic discovery.
\newblock \emph{Cell}, 180\penalty0 (4):\penalty0 688--702, 2020.

\bibitem[Stretcu et~al.(2019)Stretcu, Viswanathan, Movshovitz-Attias,
  Platanios, Ravi, and Tomkins]{stretcu2019graph}
Otilia Stretcu, Krishnamurthy Viswanathan, Dana Movshovitz-Attias, Anthony
  Platanios, Sujith Ravi, and Andrew Tomkins.
\newblock Graph agreement models for semi-supervised learning.
\newblock 2019.

\bibitem[Thakoor et~al.(2021)Thakoor, Tallec, Azar, Munos,
  Veli{\v{c}}kovi{\'c}, and Valko]{thakoor2021bootstrapped}
Shantanu Thakoor, Corentin Tallec, Mohammad~Gheshlaghi Azar, R{\'e}mi Munos,
  Petar Veli{\v{c}}kovi{\'c}, and Michal Valko.
\newblock Bootstrapped representation learning on graphs.
\newblock \emph{arXiv preprint arXiv:2102.06514}, 2021.

\bibitem[Vaswani et~al.(2017)Vaswani, Shazeer, Parmar, Uszkoreit, Jones, Gomez,
  Kaiser, and Polosukhin]{vaswani2017attention}
Ashish Vaswani, Noam Shazeer, Niki Parmar, Jakob Uszkoreit, Llion Jones,
  Aidan~N Gomez, Lukasz Kaiser, and Illia Polosukhin.
\newblock Attention is all you need.
\newblock \emph{arXiv preprint arXiv:1706.03762}, 2017.

\bibitem[Veli{\v{c}}kovi{\'c} et~al.(2017)Veli{\v{c}}kovi{\'c}, Cucurull,
  Casanova, Romero, Lio, and Bengio]{velivckovic2017graph}
Petar Veli{\v{c}}kovi{\'c}, Guillem Cucurull, Arantxa Casanova, Adriana Romero,
  Pietro Lio, and Yoshua Bengio.
\newblock Graph attention networks.
\newblock \emph{arXiv preprint arXiv:1710.10903}, 2017.

\bibitem[Veli{\v{c}}kovi{\'c} et~al.(2018)Veli{\v{c}}kovi{\'c}, Fedus,
  Hamilton, Li{\`o}, Bengio, and Hjelm]{velivckovic2018deep}
Petar Veli{\v{c}}kovi{\'c}, William Fedus, William~L Hamilton, Pietro Li{\`o},
  Yoshua Bengio, and R~Devon Hjelm.
\newblock Deep graph infomax.
\newblock \emph{arXiv preprint arXiv:1809.10341}, 2018.

\bibitem[Veli{\v{c}}kovi{\'c} et~al.(2019)Veli{\v{c}}kovi{\'c}, Ying, Padovano,
  Hadsell, and Blundell]{velivckovic2019neural}
Petar Veli{\v{c}}kovi{\'c}, Rex Ying, Matilde Padovano, Raia Hadsell, and
  Charles Blundell.
\newblock Neural execution of graph algorithms.
\newblock \emph{arXiv preprint arXiv:1910.10593}, 2019.

\bibitem[Wang et~al.(2020{\natexlab{a}})Wang, Shen, Huang, Wu, Dong, and
  Kanakia]{wang2020microsoft}
Kuansan Wang, Zhihong Shen, Chiyuan Huang, Chieh-Han Wu, Yuxiao Dong, and
  Anshul Kanakia.
\newblock Microsoft academic graph: When experts are not enough.
\newblock \emph{Quantitative Science Studies}, 1\penalty0 (1):\penalty0
  396--413, 2020{\natexlab{a}}.

\bibitem[Wang et~al.(2020{\natexlab{b}})Wang, Witek, Landrum, and
  Riniker]{wang2020improving}
Shuzhe Wang, Jagna Witek, Gregory~A Landrum, and Sereina Riniker.
\newblock Improving conformer generation for small rings and macrocycles based
  on distance geometry and experimental torsional-angle preferences.
\newblock \emph{Journal of chemical information and modeling}, 60\penalty0
  (4):\penalty0 2044--2058, 2020{\natexlab{b}}.

\bibitem[Wang et~al.(2021)Wang, Jin, Zhang, Yu, Zhang, and Wipf]{wang2021bag}
Yangkun Wang, Jiarui Jin, Weinan Zhang, Yong Yu, Zheng Zhang, and David Wipf.
\newblock Bag of tricks for node classification with graph neural networks.
\newblock \emph{arXiv preprint arXiv:2103.13355}, 2021.

\bibitem[Wu et~al.(2019)Wu, Souza, Zhang, Fifty, Yu, and
  Weinberger]{wu2019simplifying}
Felix Wu, Amauri Souza, Tianyi Zhang, Christopher Fifty, Tao Yu, and Kilian
  Weinberger.
\newblock Simplifying graph convolutional networks.
\newblock In \emph{International conference on machine learning}, pages
  6861--6871. PMLR, 2019.

\bibitem[Ying et~al.(2018)Ying, He, Chen, Eksombatchai, Hamilton, and
  Leskovec]{ying2018graph}
Rex Ying, Ruining He, Kaifeng Chen, Pong Eksombatchai, William~L Hamilton, and
  Jure Leskovec.
\newblock Graph convolutional neural networks for web-scale recommender
  systems.
\newblock In \emph{Proceedings of the 24th ACM SIGKDD International Conference
  on Knowledge Discovery \& Data Mining}, pages 974--983, 2018.

\bibitem[Yu et~al.(2020)Yu, Shen, Li, and Lerer]{yu2020scalable}
Lingfan Yu, Jiajun Shen, Jinyang Li, and Adam Lerer.
\newblock Scalable graph neural networks for heterogeneous graphs.
\newblock \emph{arXiv preprint arXiv:2011.09679}, 2020.

\bibitem[Zeng et~al.(2019)Zeng, Zhou, Srivastava, Kannan, and
  Prasanna]{zeng2019graphsaint}
Hanqing Zeng, Hongkuan Zhou, Ajitesh Srivastava, Rajgopal Kannan, and Viktor
  Prasanna.
\newblock Graphsaint: Graph sampling based inductive learning method.
\newblock \emph{arXiv preprint arXiv:1907.04931}, 2019.

\bibitem[Zhu and Ghahramani(2002)]{zhu2002learning}
Xiaojin Zhu and Zoubin Ghahramani.
\newblock Learning from labeled and unlabeled data with label propagation.
\newblock 2002.

\bibitem[Zhu et~al.(2020)Zhu, Xu, Yu, Liu, Wu, and Wang]{zhu2020deep}
Yanqiao Zhu, Yichen Xu, Feng Yu, Qiang Liu, Shu Wu, and Liang Wang.
\newblock Deep graph contrastive representation learning.
\newblock \emph{arXiv preprint arXiv:2006.04131}, 2020.

\bibitem[Zou et~al.(2019)Zou, Hu, Wang, Jiang, Sun, and Gu]{zou2019layer}
Difan Zou, Ziniu Hu, Yewen Wang, Song Jiang, Yizhou Sun, and Quanquan Gu.
\newblock Layer-dependent importance sampling for training deep and large graph
  convolutional networks.
\newblock \emph{arXiv preprint arXiv:1911.07323}, 2019.

\end{thebibliography}

\end{document}